\documentclass[manuscript,screen,sigconf,nonacm,natbib=false]{acmart}
\renewcommand
\footnotetextcopyrightpermission[1]{} 
\makeatletter
\renewcommand\@formatdoi[1]{\ignorespaces}
\makeatother

\AtBeginDocument{%
  \providecommand\BibTeX{{%
    \normalfont B\kern-0.5em{\scshape i\kern-0.25em b}\kern-0.8em\TeX}}}
\usepackage[numbers]{natbib}
\usepackage{caption}
\usepackage{subcaption}
\usepackage{makecell}

\setcopyright{none}

\begin{document}

\title{Outlier detection in maritime environments using AIS data and deep recurrent architectures}

\author{Constantine Maganaris}
\affiliation{%
  \institution{National Technical University of Athens, School of Rural, Surveying and Geoinformatics Engineering}
  \city{Athens}
  \country{Greece}}
\email{c@maganaris.com}
\authornote{All authors contributed equally to this research.}
\orcid{0000-0003-2738-684X}
\author{Eftychios Protopapadakis}
\authornotemark[1]
\email{eftprot@uom.edu.gr}
\affiliation{%
  \institution{University of Macedonia, Department of Applied Informatics}
  \city{Thessaloniki}
  \country{Greece}
}
 
 \author{Nikolaos Doulamis}
\affiliation{%
  \institution{Institute of Communications and Computer Systems}
  \city{Athens}
  \country{Greece}}
\email{ndoulam@cs.ntua.gr}

\renewcommand{\shortauthors}{C. Maganaris, E. Protopapadakis, N. Doulamis}

%

\begin{abstract}

A methodology based on deep recurrent models for maritime surveillance, over publicly available Automatic Identification System (AIS) data, is presented in this paper. The setup employs a deep Recurrent Neural Network (RNN)-based model, for encoding and reconstructing the observed ships' motion patterns. Our approach is based on a thresholding mechanism, over the calculated errors between observed and reconstructed motion patterns of maritime vessels. Specifically, a deep-learning framework, i.e. an encoder-decoder architecture, is trained using the observed motion patterns, enabling the models to learn and predict the expected trajectory, which will be compared to the effective ones. Our models, particularly the bidirectional GRU with recurrent dropouts, showcased superior performance in capturing the temporal dynamics of maritime data, illustrating the potential of deep learning to enhance maritime surveillance capabilities. Our work lays a solid foundation for future research in this domain, highlighting a path toward improved maritime safety through the innovative application of technology. 
\end{abstract}

\keywords{datasets, neural networks, segmentation, RNN, GRU, deep learning, AI, AIS, maritime, outlier detection}

\maketitle
\section{Introduction}
In the maritime domain the creation and utilization of reliable mechanisms for anomaly detection is crucial \cite{soldi2021space1}.  Nowadays, the term anomaly detection employs much more than just deviations from a predifined set of rules. Operators are interested in early indications of potential risks, such as environmental hazards, navigational errors, or even covert suspicious activities. These subtle changes in trajectories, if promptly detected, could significantly bolster maritime security, aid in risk mitigation, and provide timely insights to law enforcement authorities. As such, the utilization of deep learning approaches, based on various data sources, is an active research field \cite{soldi2021space2}.  

The Automatic Identification System (AIS) is a widely used tracking system in the maritime industry that enables vessels to automatically broadcast information about their identity, position, speed, and other relevant information \cite{ma2024big}. AIS was originally developed to enhance maritime safety and navigation, but it also serves other purposes such as vessel traffic management, maritime domain awareness, and search and rescue operations. As such, it serves as an extremely useful data source for any analysis that involves outliers detection.

The amount of data provided by AIS dictates the utilization of deep learning approaches, which are capable of capturing complex dynamics in maritime environments \cite{protopapadakis2017stacked}. Furthermore, since we are handling sequences of information, i.e. ships motion patterns as time series, RNN architectures should be considered. In this study, we employ a deep-learning RNN-based approach for encoding and reconstructing the observed ships' motion patterns. The outlier detection mechanisms are based on the reconstruction errors between the model and the observed values.

\begin{figure*}[htb!]
     \centering
         \centering
         \includegraphics[width=0.97\textwidth]{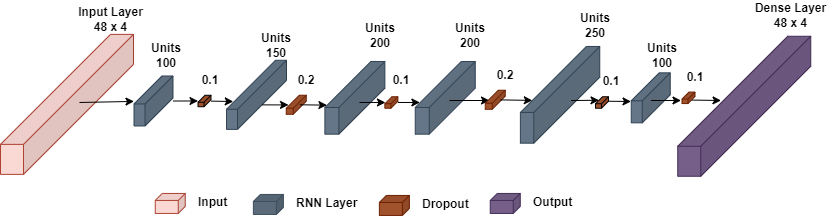}
         \caption{Schematic of Simple RNN, GRU models architecture without recurrent dropouts nor bidirectional layers.}
         \label{fig:RNN_simple}
\end{figure*}

\section{Related work}
It's essential to contextualize our approach within the broader landscape of trajectory outlier detection methods, especially those applied to maritime navigation. In particular, a distinctive methodology, as presented by Shuang et al.\cite{Shuang_2020}, stands out for its innovative use of mathematical formulas rather than deep learning techniques. The TODDT algorithm capitalizes on dynamic differential thresholds, significantly enhancing outlier detection by incorporating factors such as speed and course over ground (COG). Our methodology emphasizes characteristics like speed and COG, underscoring the relevance of these factors in achieving accurate and efficient outlier detection in maritime trajectory analysis.

The research presented  by Shaoqing Guo et al. \cite{jmse9060609} introduces an innovative anomaly detection method for AIS  trajectory, utilizing kinematic interpolation. This approach makes comprehensive use of the kinematic information of the ship in the AIS data. The method employs three steps to obtain non-error AIS trajectories: (1) data preprocessing, (2) kinematic estimation, and (3) error clustering. It should be noted that steps (2) and (3) are involved in an iterative process to determine all abnormal data.

The work of Singh et al. \cite{9109806} presents an anomaly detection framework, based on a multi-class artificial neural network (ANN), for the distinction between intentional and unintentional AIS on-off switching anomalies. Leveraging real-world AIS data, including position, speed, course, and timing, the framework accurately identifies message dropouts caused by channel effects or deliberate actions, achieving impressive accuracy. The framework also allows for further classification of anomalies, such as those occurring during mooring, U-turns, or unauthorized entry into restricted zones. The primary distinction in the ANN technique employed in this study is that it utilizes supervised learning, whereas ours is based on unsupervised learning.

Another proposed methodology described in "Toward a Deep Learning Approach to Behavior-based AIS Traffic Anomaly Detection"\cite{blauwkamp2018toward}, the authors intriguingly propose the use of images for enhancing anomaly detection, presenting a suggestion for future research. While not integrated into the current framework, this approach suggests leveraging visual data of trajectories to potentially uncover patterns and anomalies. This prospective method opens avenues for richer multidimensional analysis, promising to enhance detection capabilities by incorporating visual cues alongside existing parameters.

Finally, the RANGER project \cite{Karagiannidis2019RANGERRA} research, outlier detection is achieved through an integrated approach utilizing AIS, Over-the-Horizon, and Photonics Enhanced Multiple-Input Multiple-Output radar data. This method employs stacked autoencoders for dimensionality reduction and density-based clustering for the identification of anomalous vessel behavior. The combination of multiple data sources enhances maritime surveillance capabilities but requires access in advanced hardware, not accessible to the general audience, and complicated trajectories merging algorithms.

\subsection{Our contribution}



In this paper, we present an innovative approach diverging, which does not rely on specialized mathematical concepts, nor on successful projection in lower dimensional spaces, does not need labeled datasets, and is not based on data sourced from restricted channels, such as satellite feeds or advanced radar systems. In particular the adopted scheme takes as inputs AIS data and employs unsupervised, deep-learning, RNN-based architectures for the identification of potential outliers. Additionally, the current setup allows for an easy parameterization and utilization with different data, if they become available.

\begin{figure*}[h!]
     \centering
         \centering
         \includegraphics[width=0.85\textwidth]{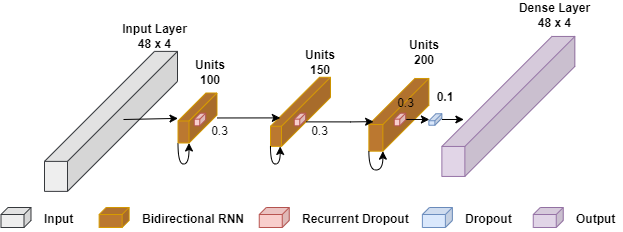}
         \caption{Schematic of SimpleRNN, GRU models architecture with recurrent dropouts in bidirectional layers.}
         \label{fig:RNN_bidirectional}
\end{figure*}

\section{Proposed Methodology}

The employed outlier detection methodology hinges upon the utilization of a thresholding mechanism, over the calculated errors between observed and reconstructed motion patterns of maritime vessels. Specifically, a deep learning framework, i.e. an RNN-based encoder-decoder architecture, is trained using the observed motion patterns. Subsequently, each new observation corresponding to the motion patterns of a ship within a predetermined temporal window, is fed into the trained model. The output is compared against the observed values, providing an error score. Given a set of $n$ observations, representing ships' trajectories, at time $t$, the higher reconstruction error indicate potential outliers.

\subsection{Input setup}
Our input sequences for the deep learning models consist of longitudinal (LON) and latitudinal (LAT) coordinates, as well as Speed Over Ground (SOG) and Course Over Ground (COG) values extracted from Automatic Identification System (AIS) data. By incorporating these parameters into our input sequences, we capture the spatiotemporal dynamics of vessel movements with high granularity, enabling the models to learn and predict the expected trajectory, which will be compared to the effective ones. The behaviour of a ship is described through a tensor of the form $t \times f$, where $t$ denotes the number of previous observations for the ships values, over a predefined time steps, and $f$ stands for the feature values, i.e. LAT, LOG, SOG, and COG, respectively. 

\subsection{Evaluated Models' Architectures}
Deep learning models have been evaluated, over the provided dataset. These models employed two different types of Recurrent Neural Networks (RNN), specifically Simple RNN and Gated Recurrent Unit \cite{cho2014properties} (GRU). Each model was constructed with an input layer designed to handle 48 timesteps, each comprising 4 parameters (i.e. LAT, LON, SOG, COG). The RNN layers in our models were interspersed with different dropout layers to prevent overfitting, and concluded with an \textbf{output layer, mirroring the input layer} to predict the same parameters accurately and not future ones.

Similar features across all models was the utilization of the Tanh activation function (since the missing entries are represented by the value -1), the Adam optimization function, and the mean squared error loss function. These choices were depicted by the nature of our data and the need for a robust framework capable of capturing and predicting the temporal dynamics of maritime traffic with high precision after doing numerous examples.

We adopted two distinct approaches for each model to explore the impact of architecture variations on performance. The initial approach, as illustrated in Fig. \ref{fig:RNN_simple}, incorporated separate dropout layers within the Model's architecture but it has greater depth in model's layers and was trained for more epochs (see Table \ref{table:training_summary_table}). The conventional dropout technique selected here, involves randomly omitting a fraction of the network's neurons during each training iteration to enhance generalization and mitigate the risk of overfitting.

In the second approach, depicted in Fig. \ref{fig:RNN_bidirectional}, we innovated by integrating Bidirectional RNN layers accompanied by Recurrent dropout mechanisms, a strategy inspired by Yarin Gal's thesis \cite{Gal2016UncertaintyDL}. This method differs from the traditional dropout by maintaining a consistent set of units dropped across all timesteps, as opposed to random selection at each step. Such a strategy ensures the preservation of temporal relationships within the sequence data, thereby enabling the network to effectively learn and memorize information throughout the training process without the disruption typically caused by variable dropout patterns.

\section{Experimental setup}
The Models and the data transformation scripts, were developed in Python 3 using TensorFlow and Keras libraries. Models were trained on a PC using Windows 10 OS with 16 core CPU and
64GB RAM.

\subsection{Dataset description}

The data set \cite{NOAA} utilized in this investigation was acquired from MarineCadastre.gov, which provides data from the AIS derived from the U.S. Coast Guard's Nationwide Automatic Identification System (NAIS). These data are publicly available and cover U.S. coastal waters for calendar years 2009 through 2024. The data set mainly covers maritime activities within the coordinates of -168° W to -60° E and 50° N to 15° S. However, our findings reveal the presence of vessels operating outside these geographical boundaries, indicating a broader scope of maritime traffic than initially anticipated.

For our study, we selected a subset of data spanning 100 days, from March 6, 2019, to June 13, 2019. Our analysis focused on several key fields from the AIS data, namely Maritime Mobile Service Identity (MMSI), BaseDateTime, Latitude (LAT), Longitude (LON), Speed Over Ground (SOG), Course Over Ground (COG) and Length of the vessel.

\begin{figure*}[!ht]
     \centering
         \centering
         \includegraphics[width=0.75\textwidth]{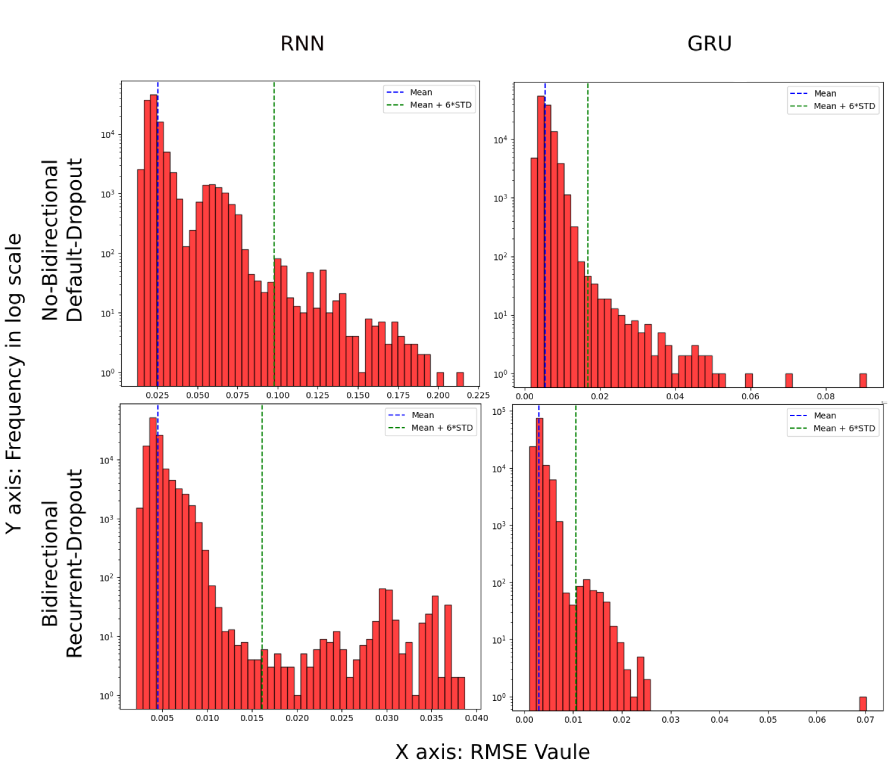}
         \caption{Root Mean Squared Error Histograms of Test dataset}
         \label{fig:RMSE_Histogram}
\end{figure*}

\subsection*{Data Filtering and Preparation}

The initial step in data preparation involved filtering the data set to include only vessels longer than 20 meters, with the aim of focusing on significant maritime traffic. The AIS data, which are daily records for each U.S. coastal water zone, required aggregation and sorting by the BaseDateTime field for each vessel identified by its MMSI. This step ensured chronological order, addressing unsorted entries present in the original dataset. 

Subsequent filtering was designed to standardize the temporal resolution of the data. We retained AIS entries with time intervals of 30 minutes starting at 00:00 each day, ensuring a uniform data set conducive to time-series analysis. This process also included filtering out vessels with fewer than 20 entries per day to maintain a robust data set for each vessel, with the expectation of 48 entries per vessel per day.
Additionally, linear interpolation was employed for each vessel's daily data to handle missing data points within the observed range, with a limit of up to 20 entries per vessel to ensure data reliability. This approach helps to exclude extrapolation beyond the data set's bounds. It's important to note that while this method aids in filling gaps in the data, it may result in some interpolated points being inaccurately positioned in land areas on the map. This is particularly relevant when vessels navigate within river channels where their courses might lead them close to or across terrestrial boundaries.

The processed data were structured into vectors representing each vessel's daily activity over the selected 100-day period, with a final format of 48 entries per day for each of the four key attributes (LAT, LON, SOG, COG). 
\begin{equation} f_{t_i} = \{\text{LAT}, \text{LON}, \text{COG}, \text{SOG}\}\end{equation}
Since the final vectors included 100 days data for each vessel, there were vectors that included only NaN values. To address the presence of NaN values, a final filtering step was implemented. Entries with more than 30\% missing values were identified and excluded from the data set. The remaining NaN values were substituted with a placeholder value of -1.

Before inputting the data into our deep learning models, a crucial preprocessing step involved normalizing the variables LAT, LON, COG and SOG to ensure that all input features have a similar scale. Normalization for each variable was performed using the following formula:

\begin{equation}
    x_{normalized} = \frac{x - \text{global\_min}[\text{'Var'}]}{\text{global\_max}[\text{'Var'}] - \text{global\_min}[\text{'Var'}]}
\end{equation}

Where $x$ represents the original value of the variable, $\text{global\_min}[\text{'Var'}]$ and $\text{global\_max}[\text{'Var'}]$ are the global minimum and maximum values of the variable, respectively, across the entire dataset. This formula was applied individually to each of the variables: LAT, LON, COG, and SOG, resulting in normalized values that range between 0 and 1 (excluding the missing values which were replaced by -1). Such normalization is critical for maintaining consistency in the feature space and ensuring that the model's performance is not adversely affected by the range of input values.

In our methodology, all models were trained and evaluated using the same dataset (i.e. Table \ref{tab:dataset_summary}), initially containing 1,474,100 entries with dimensions (48, 5). We introduced an additional variable (MMSI), to uniquely identify each vessel, though this variable was not used in the training process. Our data set was segmented into training vectors with dimensions (472204, 48, 4), a validation set comprising 20\% of the training set and a test set sized (118052, 48, 4). In all sets an MMSI parameter was kept to be able to identify each vessel on the analysis of the results.

\begin{table}[!htb]
\centering
\caption{Summary of Dataset Size}
\label{tab:dataset_summary}
\begin{tabular}{@{}lccc@{}}
\toprule
Dataset         & Num of instances & Observation shape   \\ \midrule
Pre-cleansing Set & 1,474,100 & (48, 5) \\
Train Set & 472,204 & (48, 4)  \\
Validation Set  & {20\% of Train Instances} & (48, 4) \\
Test Set & 118,052 & (48, 4)  \\ \bottomrule
\end{tabular}
\end{table}

%

\subsection{Models performance}
The bidirectional concept we employed involves processing the data in both forward and reverse directions. This dual-path architecture allows the network to capture dependencies not just from past to future (as is typical in unidirectional RNNs) but also from future to past. This approach enriches the model's context understanding, as is clear also from Fig. \ref{fig:RMSE_Histogram} where we can see that the RMSE values of Reccurent-Droput models have much less deviation from the mean RMSE value compared to their No-Bidirectional implementations.
Finally in this implementation we also added a separate dropout to regularize the Dense layer (see  Fig. \ref{fig:RNN_bidirectional}). 

\begin{figure*}[!h]
     \centering
     \begin{subfigure}[b]{0.45\textwidth}
         \centering
         \includegraphics[width=1.0\textwidth]{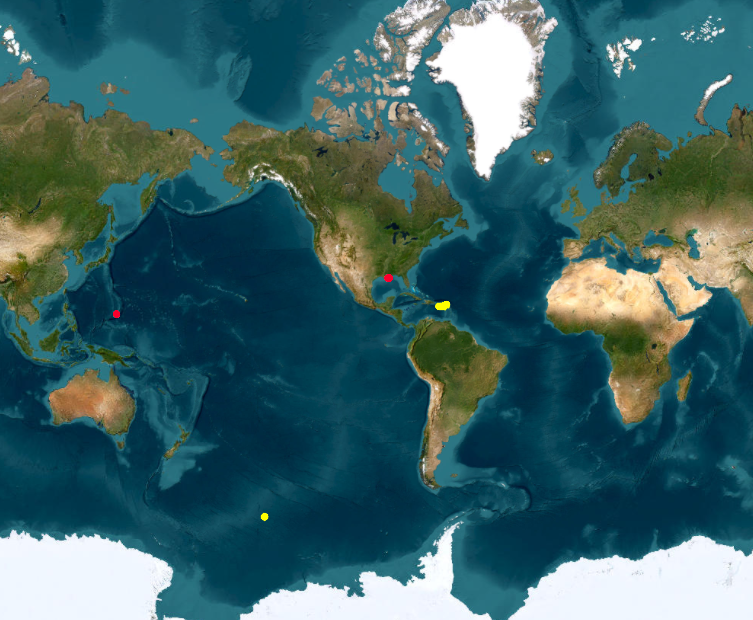}
         \caption{}
    \end{subfigure}
\hfill
        \begin{subfigure}[b]{0.463\textwidth}
         \centering
         \includegraphics[width=1.0\textwidth]{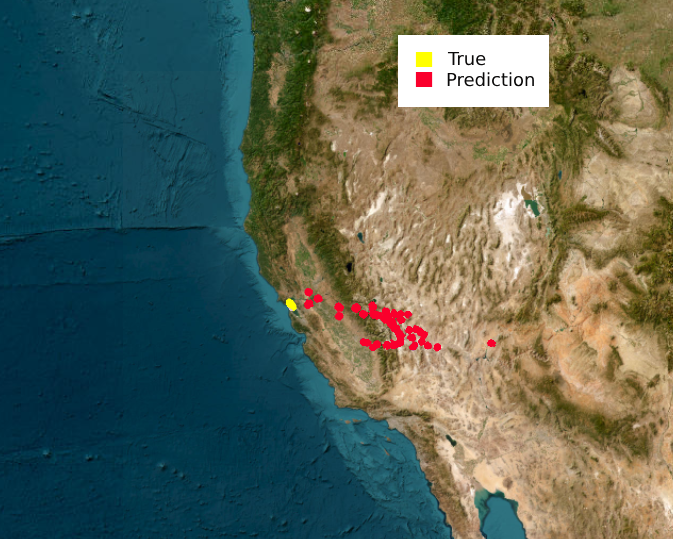}
         \caption{}
    \end{subfigure}
           \caption{Outliers from Test data using the Bidirectional GRU model}
         \label{fig:GT_vs_pred_timeseries_LAT_LON_GRU_bidirectional_MAP}
\end{figure*}

\begin{figure*}[!h]
     \centering
         \centering
         \includegraphics[width=1.0\textwidth]{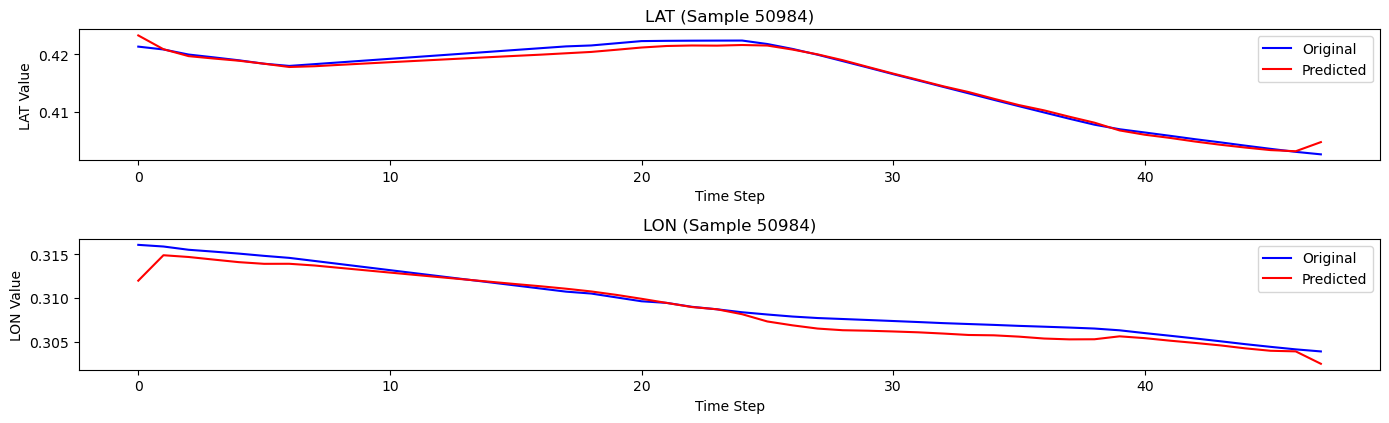}
         \caption{Ground Truth vs Prediction in a randomly selected vessel from Test data, for Latitude and Longtitude in a day's data using the Bidirectional GRU model}
         \label{fig:GT_vs_pred_timeseries_LAT_LON_GRU_bidirectional}
\end{figure*}

In our approach to train the deep learning models, a deliberate decision was made to limit the extent of training to avoid that the models acquire too much knowledge of the outliers. This strategic choice ensured that the models could generally predict regular patterns without becoming too tuned to outliers. Thus overfitting was meticulously avoided to ensure that the models generalize well on new data and be effective in identifying genuine anomalies, rather than memorizing specific outlier instances. This careful balancing act allowed us to maintain the integrity of our research objectives, focusing on the detection and analysis of navigational anomalies without compromising the models' ability to generalize across the dataset, which is easily understandable from the histograms of RMSEs \ref{fig:RMSE_Histogram} in the Test data.

Of course the SimpleRNN models, despite its theoretical capacity to retain information across numerous timesteps, is often too elementary for practical applications and this is also observable in our results (see Fig. \ref{fig:RMSE_Histogram}). This stems from its inherent inability to effectively learn long-term dependencies, a phenomenon largely attributed to the vanishing gradient problem as identified by Hochreiter et. al. in "Learning long-term dependencies with gradient descent is difficult" \cite{BengioSimplisticRNN}.
\begin{table*}[h!]
\centering
\begin{tabular}{p{0.2\linewidth}p{0.17\linewidth}p{0.2\linewidth}p{0.18\linewidth}p{0.1\linewidth}}
\hline
\textbf{Model} & \textbf{Training Time (seconds)} & \textbf{Loss (Last Epoch)} & \textbf{Validation Loss (Last Epoch)} & \textbf{Epochs} \\
\hline
GRU & 47349 & $9.90 \times 10^{-5}$ & $3.31 \times 10^{-5}$ & 10\\
Bidirectional GRU & 61259 & $1.88 \times 10^{-5}$ & $2.28 \times 10^{-5}$ & 5 \\
Simple RNN & 12778 & $0.0020$ & $7.89 \times 10^{-4}$ & 10\\
Bidirectional & 8652 & $5.65 \times 10^{-5}$ & $2.41 \times 10^{-5}$ & 5 \\Simple RNN \\
\hline
\end{tabular}
\caption{Model Training parameters and metadata.}
\label{table:training_summary_table}
\end{table*}
From the results presented in Table \ref{table:training_summary_table}, it is evident that the bidirectional models with recurrent dropouts outperform their simpler counterparts, despite having less depth and fewer epochs, as reflected in their lower test loss and validation loss values. This is also evident from the RMSEs histograms of the Test data in Fig. \ref{fig:RMSE_Histogram}. From the data in Table \ref{table:training_summary_table} it is obvious that most bidirectional versions of the models required significantly more training time than their unidirectional counterparts, despite having shallower network depths and epochs. This phenomenon highlights the complexity and computational demands of processing data in both forward and reverse directions.

\subsection{Experimental results}

Following the training of the various model architectures discussed previously, we proceeded to predict the time-series data from the Test set. We calculated the Root Mean Squared Errors (RMSEs) by comparing each ground truth vector with its corresponding predicted vector. To identify potential outliers, we considered vectors of vessel trajectories that significantly deviated from the mean RMSE of each model, specifically adopting a threshold of six standard deviations from the mean RMSE (see Fig. \ref{fig:RMSE_Histogram}). This criterion was chosen based on the observation that the RMSE distributions are right-skewed with positive values, making the six standard deviation interval a robust marker for outlier detection. Nevertheless, it is essential to mention that the Simple RNN models exhibited more significant differences in RMSE values and showed inferior performance in forecasting vessel trajectories, consequently, we do not consider them as an acceptable methodology for outlier detection.

Among all models, the bidirectional GRU model exhibited the least deviation from the mean RMSE, making it our preferred model. By examining the results in the best RMSE scores we managed to identify numerous vessels that generally followed a normal course and then were detected transmitting signals from entirely different locations within the day. This anomaly is visually represented in Fig. \ref{fig:GT_vs_pred_timeseries_LAT_LON_GRU_bidirectional_MAP} (a) , where each color represents the stigmas of the vessel within the data of one day.

Although the bidirectional GRU model exhibited significant prediction results in many cases (i.e., Fig. \ref{fig:GT_vs_pred_timeseries_LAT_LON_GRU_bidirectional}), it did not achieve perfect accuracy in every case of predicting outliers, as you can see in Fig. \ref{fig:GT_vs_pred_timeseries_LAT_LON_GRU_bidirectional_MAP} (b) the model cannot accurately predict all vessels' courses to such an extent, so they will have such a low RMSE. 
We should note here that the predictions also consider the SOG and COG values. As such the LON and LAT estimation are prone to variations, induced by the SOG and COG variations. Such cases result in over-complicated trajectory predictions, as the one in Fig. \ref{fig:GT_vs_pred_timeseries_LAT_LON_GRU_bidirectional_MAP} (b).

\begin{figure*}[!h]
     \centering
         \centering
         \includegraphics[width=1.0\textwidth]{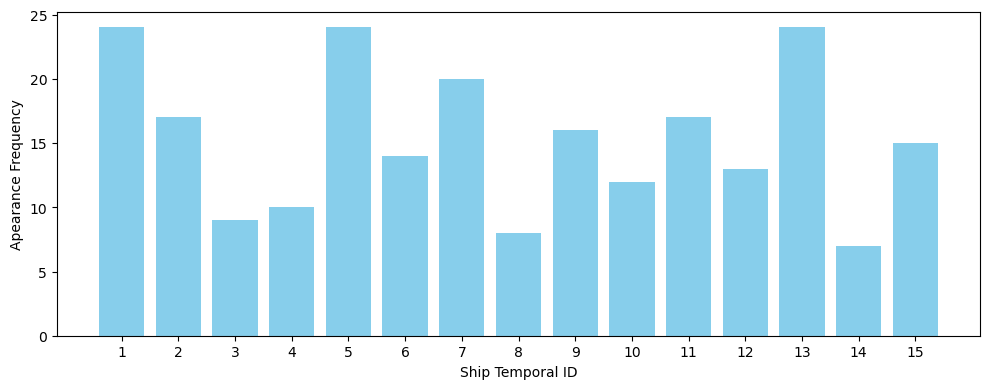}
         \caption{Unique MMSI Counts greater than $\bar{RMSE} + 6 \times \sigma$  threshold, in bidirectional GRU}
         \label{fig:high_frequencies_RMSE_counts}
\end{figure*}

Further analysis of the most significant outliers, those exceeding the mean RMSE by more than six standard deviations, allowed us to track how frequently each vessel appeared in this subset of data. Through this analysis, depicted in Fig. \ref{fig:high_frequencies_RMSE_counts}, we observed multiple vessels (using numbers instead of their true MMSIs) repeatedly flagged as outliers, indicating potential equipment malfunctions or suspicious behavior. Setting a threshold of five appearances in this set, we could confidently identify vessels exhibiting consistently anomalous patterns within 100 days data. This additional layer of analysis reinforces our methodology's effectiveness in detecting vessels with unusual behavior or technical issues, highlighting our model's utility in maritime surveillance and security.

\section{Dataset limitations}
Despite the invaluable insights AIS data provide into maritime traffic and vessel behavior, several limitations merit attention to fully understand the context and potential biases within our analysis.

Firstly, the dataset is not exhaustive of all maritime traffic, as not all vessels consistently transmit AIS signals. The likelihood of AIS data transmission can vary depending on the vessel's location and the coverage of the AIS receiving stations. This results in missing entries for certain regions or periods, introducing gaps in the dataset that could affect the comprehensiveness of our analysis.

\begin{figure*}[!h]
\hfill
\begin{subfigure}[b]{0.47\textwidth}
     \centering
         \includegraphics[width=1.0\textwidth]{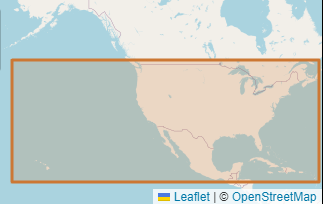}
         \caption{AIS coverage Area}
         \label{fig:AIS_Area_coverage}
    \end{subfigure}
\hfill
     \begin{subfigure}[b]{0.47\textwidth}
         \centering
         \includegraphics[width=0.78\textwidth]{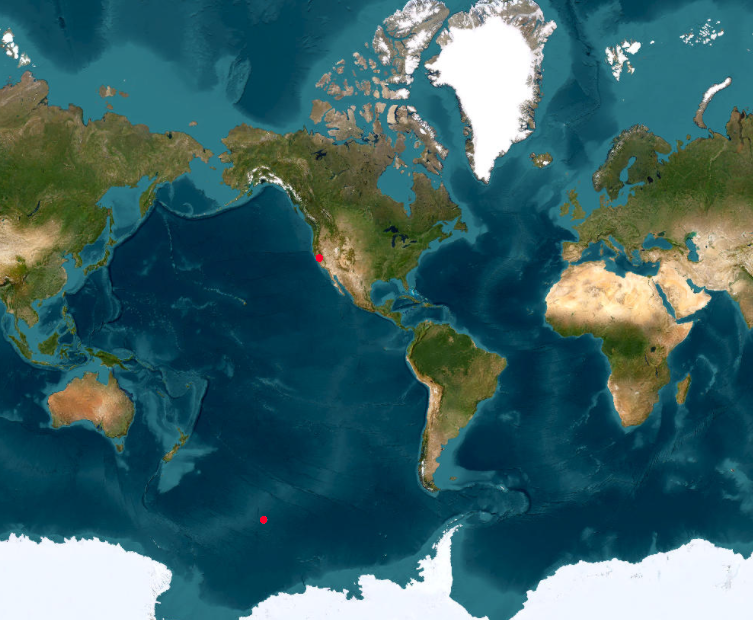}
         \caption{Vessel outside coverage area}
         \label{fig:Ship_outside_coverage}
    \end{subfigure}
           \caption{}
\hfill
\end{figure*}

Furthermore, an intriguing observation from our dataset is the presence of AIS signals transmitted from vessels at considerable distances, such as those from the Southern Ocean (i.e. Fig. \ref{fig:Ship_outside_coverage}) which is outside the coverage area noted by NOAA (i.e. Fig. \ref{fig:AIS_Area_coverage}), whereas vessels within closer proximities, like those navigating the Gulf of Mexico, may not always transmit AIS data. This variability in the availability of data prompts an investigation into the dynamics affecting the transmission and reception of AIS signals, which could be attributed to either technical challenges or deliberate decisions by those managing the vessels. Such signals also give the opportunity for further research by analyzing historical data of these ships in greater time-length to gain more valuable insights for potential suspicious activities.

To address missing data within our dataset, we employed linear interpolation to estimate the values of gaps in the data record. Although this approach allows us to maintain a continuous dataset for analysis, it is not without limitations. Specifically, interpolated points may not accurately represent the true position of the vessel, particularly for vessels that navigate close to land or within river channels. This could lead to incorrect placement of vessels in land areas within the map, based on the interpolated data.

\section{Conclusions}

In conclusion, our study advances the field of maritime navigation anomaly detection by leveraging deep learning models to analyze AIS data from MarineCadastre. Through a meticulous experimental setup employing RNNs, including SimpleRNN and GRU, we have demonstrated the effectiveness of different architectures in identifying navigational anomalies. Our models, particularly the bidirectional GRU with recurrent dropouts, showcased superior performance in capturing the temporal dynamics of maritime data, illustrating the potential of deep learning to enhance maritime surveillance.

Our deliberate approach to model training, ensuring not to over-fit to outliers, has underscored the importance of a balanced methodology that prioritizes the detection of genuine anomalies over memorizing data idiosyncrasies. This strategy, combined with our thorough analysis, has notably identified certain vessels that frequently appear as outliers, underscoring the recurrent nature of specific anomalies within the maritime domain.

Moving forward, the integration of sophisticated deep learning techniques with an exhaustive analysis of AIS data is poised to significantly enhance maritime surveillance capabilities. Our work lays a solid foundation for future research in this domain, highlighting a path toward improved maritime safety through the innovative application of technology. This research marks an important milestone in the evolution of anomaly detection in maritime navigation, paving the way for a safer and more secure maritime environment.

\begin{acks}
This paper is supported by the project with ID 95030400 "Photogrammetry \& Geoinformatics" of the National Technical University of Athens.
\end{acks}

\bibliographystyle{ACM-Reference-Format}
\bibliography{AIS_outlier_detection}


\begin{thebibliography}{13}


\ifx \showCODEN    \undefined \def \showCODEN     #1{\unskip}     \fi
\ifx \showDOI      \undefined \def \showDOI       #1{#1}\fi
\ifx \showISBNx    \undefined \def \showISBNx     #1{\unskip}     \fi
\ifx \showISBNxiii \undefined \def \showISBNxiii  #1{\unskip}     \fi
\ifx \showISSN     \undefined \def \showISSN      #1{\unskip}     \fi
\ifx \showLCCN     \undefined \def \showLCCN      #1{\unskip}     \fi
\ifx \shownote     \undefined \def \shownote      #1{#1}          \fi
\ifx \showarticletitle \undefined \def \showarticletitle #1{#1}   \fi
\ifx \showURL      \undefined \def \showURL       {\relax}        \fi
\providecommand\bibfield[2]{#2}
\providecommand\bibinfo[2]{#2}
\providecommand\natexlab[1]{#1}
\providecommand\showeprint[2][]{arXiv:#2}

\bibitem[Bengio et~al\mbox{.}(1994)]%
        {BengioSimplisticRNN}
\bibfield{author}{\bibinfo{person}{Y. Bengio}, \bibinfo{person}{P. Simard},
  {and} \bibinfo{person}{P. Frasconi}.} \bibinfo{year}{1994}\natexlab{}.
\newblock \showarticletitle{Learning long-term dependencies with gradient
  descent is difficult}.
\newblock \bibinfo{journal}{\emph{IEEE Transactions on Neural Networks}}
  \bibinfo{volume}{5}, \bibinfo{number}{2} (\bibinfo{year}{1994}),
  \bibinfo{pages}{157--166}.
\newblock
\urldef\tempurl%
\url{https://doi.org/10.1109/72.279181}
\showDOI{\tempurl}


\bibitem[Blauwkamp et~al\mbox{.}(2018)]%
        {blauwkamp2018toward}
\bibfield{author}{\bibinfo{person}{Daniel Blauwkamp}, \bibinfo{person}{Thuy~D
  Nguyen}, {and} \bibinfo{person}{Geoffrey~G Xie}.}
  \bibinfo{year}{2018}\natexlab{}.
\newblock \showarticletitle{Toward a deep learning approach to behavior-based
  AIS traffic anomaly detection}. In \bibinfo{booktitle}{\emph{Dynamic and
  Novel Advances in Machine Learning and Intelligent Cyber Security (DYNAMICS)
  Workshop, San Juan, PR}}.
\newblock


\bibitem[Cho et~al\mbox{.}(2014)]%
        {cho2014properties}
\bibfield{author}{\bibinfo{person}{Kyunghyun Cho}, \bibinfo{person}{Bart van
  Merrienboer}, \bibinfo{person}{Dzmitry Bahdanau}, {and}
  \bibinfo{person}{Yoshua Bengio}.} \bibinfo{year}{2014}\natexlab{}.
\newblock \bibinfo{title}{On the Properties of Neural Machine Translation:
  Encoder-Decoder Approaches}.
\newblock
\newblock
\showeprint[arxiv]{1409.1259}~[cs.CL]


\bibitem[Gal(2016)]%
        {Gal2016UncertaintyDL}
\bibfield{author}{\bibinfo{person}{Yarin Gal}.}
  \bibinfo{year}{2016}\natexlab{}.
\newblock \emph{\bibinfo{title}{Uncertainty in Deep Learning}}.
\newblock \bibinfo{thesistype}{Ph.\,D. Dissertation}.
  \bibinfo{school}{University of Cambridge}.
\newblock


\bibitem[Guo et~al\mbox{.}(2021)]%
        {jmse9060609}
\bibfield{author}{\bibinfo{person}{Shaoqing Guo}, \bibinfo{person}{Junmin Mou},
  \bibinfo{person}{Linying Chen}, {and} \bibinfo{person}{Pengfei Chen}.}
  \bibinfo{year}{2021}\natexlab{}.
\newblock \showarticletitle{An Anomaly Detection Method for AIS Trajectory
  Based on Kinematic Interpolation}.
\newblock \bibinfo{journal}{\emph{Journal of Marine Science and Engineering}}
  \bibinfo{volume}{9}, \bibinfo{number}{6} (\bibinfo{year}{2021}).
\newblock
\showISSN{2077-1312}
\urldef\tempurl%
\url{https://doi.org/10.3390/jmse9060609}
\showDOI{\tempurl}


\bibitem[Karagiannidis et~al\mbox{.}(2019)]%
        {Karagiannidis2019RANGERRA}
\bibfield{author}{\bibinfo{person}{Lazaros Karagiannidis},
  \bibinfo{person}{Dimitrios Dres}, \bibinfo{person}{Eftychios~E.
  Protopapadakis}, \bibinfo{person}{Fr{\'e}d{\'e}ric Lamole},
  \bibinfo{person}{François Jacquin}, \bibinfo{person}{Gilles Rigal},
  \bibinfo{person}{Eleftherios Ouzounoglou}, \bibinfo{person}{Dimitris
  Katsaros}, \bibinfo{person}{Alexandros Karalis}, \bibinfo{person}{Luigi
  Pierno}, \bibinfo{person}{Carmelo Mastroeni}, \bibinfo{person}{Marco
  Evangelista}, \bibinfo{person}{Valeria Fontana}, \bibinfo{person}{Domenico
  Gaglione}, \bibinfo{person}{Giovanni Soldi}, \bibinfo{person}{Paolo Braca},
  \bibinfo{person}{Sari Sarlio-Siintola}, \bibinfo{person}{Evangelos Sdongos},
  {and} \bibinfo{person}{Angelos~J. Amditis}.} \bibinfo{year}{2019}\natexlab{}.
\newblock \showarticletitle{RANGER: Radars and Early Warning Technologies for
  Long Distance Maritime Surveillance}.
\newblock
\urldef\tempurl%
\url{https://api.semanticscholar.org/CorpusID:210147185}
\showURL{%
\tempurl}


\bibitem[Ma et~al\mbox{.}(2024)]%
        {ma2024big}
\bibfield{author}{\bibinfo{person}{Quandang Ma}, \bibinfo{person}{Huan Tang},
  \bibinfo{person}{Cong Liu}, \bibinfo{person}{Mingyang Zhang},
  \bibinfo{person}{Dingze Zhang}, \bibinfo{person}{Zhao Liu}, {and}
  \bibinfo{person}{Liye Zhang}.} \bibinfo{year}{2024}\natexlab{}.
\newblock \showarticletitle{A big data analytics method for the evaluation of
  maritime traffic safety using automatic identification system data}.
\newblock \bibinfo{journal}{\emph{Ocean \& Coastal Management}}
  \bibinfo{volume}{251} (\bibinfo{year}{2024}), \bibinfo{pages}{107077}.
\newblock


\bibitem[{Office for Coastal Management}(2024)]%
        {NOAA}
\bibfield{author}{\bibinfo{person}{{Office for Coastal Management}}.}
  \bibinfo{year}{2024}\natexlab{}.
\newblock \bibinfo{title}{Nationwide Automatic Identification System 2019 from
  2019-3-6 to 2019-6-13}.
\newblock
  \bibinfo{howpublished}{\url{https://www.fisheries.noaa.gov/inport/item/62733}}.
\newblock
\newblock
\shownote{NOAA National Centers for Environmental Information}.


\bibitem[Protopapadakis et~al\mbox{.}(2017)]%
        {protopapadakis2017stacked}
\bibfield{author}{\bibinfo{person}{Eftychios Protopapadakis},
  \bibinfo{person}{Athanasios Voulodimos}, \bibinfo{person}{Anastasios
  Doulamis}, \bibinfo{person}{Nikolaos Doulamis}, \bibinfo{person}{Dimitrios
  Dres}, \bibinfo{person}{Matthaios Bimpas}, {et~al\mbox{.}}}
  \bibinfo{year}{2017}\natexlab{}.
\newblock \showarticletitle{Stacked autoencoders for outlier detection in
  over-the-horizon radar signals}.
\newblock \bibinfo{journal}{\emph{Computational intelligence and neuroscience}}
   \bibinfo{volume}{2017} (\bibinfo{year}{2017}).
\newblock


\bibitem[Shuang et~al\mbox{.}(2020)]%
        {Shuang_2020}
\bibfield{author}{\bibinfo{person}{Sun Shuang}, \bibinfo{person}{Chen Yan},
  {and} \bibinfo{person}{Zhang Jinsong}.} \bibinfo{year}{2020}\natexlab{}.
\newblock \showarticletitle{Trajectory Outlier Detection Algorithm for ship AIS
  Data based on Dynamic Differential Threshold}.
\newblock \bibinfo{journal}{\emph{Journal of Physics: Conference Series}}
  \bibinfo{volume}{1437}, \bibinfo{number}{1} (\bibinfo{date}{jan}
  \bibinfo{year}{2020}), \bibinfo{pages}{012013}.
\newblock
\urldef\tempurl%
\url{https://doi.org/10.1088/1742-6596/1437/1/012013}
\showDOI{\tempurl}


\bibitem[Singh and Heymann(2020)]%
        {9109806}
\bibfield{author}{\bibinfo{person}{Sandeep~Kumar Singh} {and}
  \bibinfo{person}{Frank Heymann}.} \bibinfo{year}{2020}\natexlab{}.
\newblock \showarticletitle{Machine Learning-Assisted Anomaly Detection in
  Maritime Navigation using AIS Data}. In \bibinfo{booktitle}{\emph{2020
  IEEE/ION Position, Location and Navigation Symposium (PLANS)}}.
  \bibinfo{pages}{832--838}.
\newblock
\urldef\tempurl%
\url{https://doi.org/10.1109/PLANS46316.2020.9109806}
\showDOI{\tempurl}


\bibitem[Soldi et~al\mbox{.}(2021a)]%
        {soldi2021space1}
\bibfield{author}{\bibinfo{person}{Giovanni Soldi}, \bibinfo{person}{Domenico
  Gaglione}, \bibinfo{person}{Nicola Forti}, \bibinfo{person}{Alessio
  Di~Simone}, \bibinfo{person}{Filippo~Cristian Daffin{\`a}},
  \bibinfo{person}{Gianfausto Bottini}, \bibinfo{person}{Dino Quattrociocchi},
  \bibinfo{person}{Leonardo~M Millefiori}, \bibinfo{person}{Paolo Braca},
  \bibinfo{person}{Sandro Carniel}, {et~al\mbox{.}}}
  \bibinfo{year}{2021}\natexlab{a}.
\newblock \showarticletitle{Space-based global maritime surveillance. Part I:
  Satellite technologies}.
\newblock \bibinfo{journal}{\emph{IEEE Aerospace and Electronic Systems
  Magazine}} \bibinfo{volume}{36}, \bibinfo{number}{9} (\bibinfo{year}{2021}),
  \bibinfo{pages}{8--28}.
\newblock


\bibitem[Soldi et~al\mbox{.}(2021b)]%
        {soldi2021space2}
\bibfield{author}{\bibinfo{person}{Giovanni Soldi}, \bibinfo{person}{Domenico
  Gaglione}, \bibinfo{person}{Nicola Forti}, \bibinfo{person}{Leonardo~M
  Millefiori}, \bibinfo{person}{Paolo Braca}, \bibinfo{person}{Sandro Carniel},
  \bibinfo{person}{Alessio Di~Simone}, \bibinfo{person}{Antonio Iodice},
  \bibinfo{person}{Daniele Riccio}, \bibinfo{person}{Filippo~Cristian
  Daffin{\`a}}, {et~al\mbox{.}}} \bibinfo{year}{2021}\natexlab{b}.
\newblock \showarticletitle{Space-based global maritime surveillance. Part II:
  Artificial intelligence and data fusion techniques}.
\newblock \bibinfo{journal}{\emph{IEEE Aerospace and Electronic Systems
  Magazine}} \bibinfo{volume}{36}, \bibinfo{number}{9} (\bibinfo{year}{2021}),
  \bibinfo{pages}{30--42}.
\newblock


\end{thebibliography}

\end{document}